\documentclass[letterpaper]{article} % DO NOT CHANGE THIS
\usepackage{aaai2027}  % DO NOT CHANGE THIS
\usepackage[hyphens]{url}  % DO NOT CHANGE THIS
\usepackage{graphicx} % DO NOT CHANGE THIS
\usepackage{natbib}  % DO NOT CHANGE THIS AND DO NOT ADD ANY OPTIONS TO IT
\usepackage{caption} % DO NOT CHANGE THIS AND DO NOT ADD ANY OPTIONS TO IT
\usepackage{pifont}
\usepackage{multirow}
\usepackage{algorithm}
\usepackage{algorithmic}
\usepackage{amsfonts}
\usepackage{amsmath} 
\usepackage{newfloat}
\usepackage{listings}
\usepackage[hyphens]{url}  % DO NOT CHANGE THIS
\usepackage{graphicx} % DO NOT CHANGE THIS
\usepackage{natbib}  % DO NOT CHANGE THIS AND DO NOT ADD ANY OPTIONS TO IT
\usepackage{caption} % DO NOT CHANGE THIS AND DO NOT ADD ANY OPTIONS TO IT
\usepackage{algorithm}
\usepackage{algorithmic}

\usepackage{newfloat}
\usepackage{listings}
\DeclareCaptionStyle{ruled}{labelfont=normalfont,labelsep=colon,strut=off} % DO NOT CHANGE THIS
\floatstyle{ruled}
\newfloat{listing}{tb}{lst}{}
\floatname{listing}{Listing}

\usepackage{booktabs}

 \nocopyright
\title{Preserving  Cultural Records: A Large-Scale Benchmark and Visual-Prior Routing Network for Ancient Book Image Restoration}
\author {
    Shumin Fan\textsuperscript{\rm 1}\equalcontrib,
    Tianyu Song\textsuperscript{\rm 2}\equalcontrib,
    Hongming Chen\textsuperscript{\rm 3},
    Xiaodong Zhou\textsuperscript{\rm 4}, 
    Xiang Chen\textsuperscript{\rm 5}\corresponding
}
\affiliations {
    \textsuperscript{\rm 1}The School of Marine Engineering, Dalian Maritime University, Dalian, China\\
    \textsuperscript{\rm 2}The College of Artificial Intelligence, Dalian Maritime University, Dalian, China\\
    \textsuperscript{\rm 3}The School of Information Science
    and Technology, Dalian Maritime University, Dalian, China\\
    \textsuperscript{\rm 4} The Department of Information and Cyber Security, Shanghai Police College, Shanghai, China\\
    \textsuperscript{\rm 5}The School of Computer Science and Engineering, Nanjing University of Science and Technology, Nanjing, China
}
\begin{document}

\maketitle

\begin{abstract}
Ancient books preserve invaluable cultural heritage, yet their images often suffer from complex degradations caused by long-term ageing, environmental exposure, and physical damage. Existing ancient book image restoration benchmarks are limited in scale and rely on simplified degradation simulations, which cannot fully represent real-world degradation patterns and restrict the generalization ability of restoration models. To address these challenges, we introduce AncBench, a large-scale synthetic paired benchmark containing 100K aligned clean-degraded image pairs. AncBench models realistic composite degradations through a comprehensive degradation pipeline that considers material ageing, paper texture variation, blur, staining, water-induced geometric distortion, and local occlusion. Furthermore, we propose AncRestore, a degradation-aware visual-prior routing network for ancient book image restoration. AncRestore first learns degradation representations from degraded images and then dynamically selects and integrates suitable visual priors according to different degradation characteristics. Guided by these degradation cues, a dynamic routing mechanism combines expert-wise and region-wise routing strategies adaptively adjusts the contribution of these priors across different regions, enabling the model to recover both global appearance and local text details. Extensive experiments show that AncRestore effectively suppresses composite degradations while preserving text structures and enhancing restoration quality against other methods, highlighting its application potential for intelligent ancient book image restoration in cultural heritage preservation.
\end{abstract}

\begin{links}
    \link{Code}{https://github.com/MingTian99/AncRestore}
    
%    \link{Datasets}{https://aaai.org/example/datasets}
%    \link{Extended version}{https://aaai.org/example/extended-version}
\end{links}

\section{Introduction}

\begin{figure}[!t]
	\centering
	\includegraphics[width=0.47\textwidth]{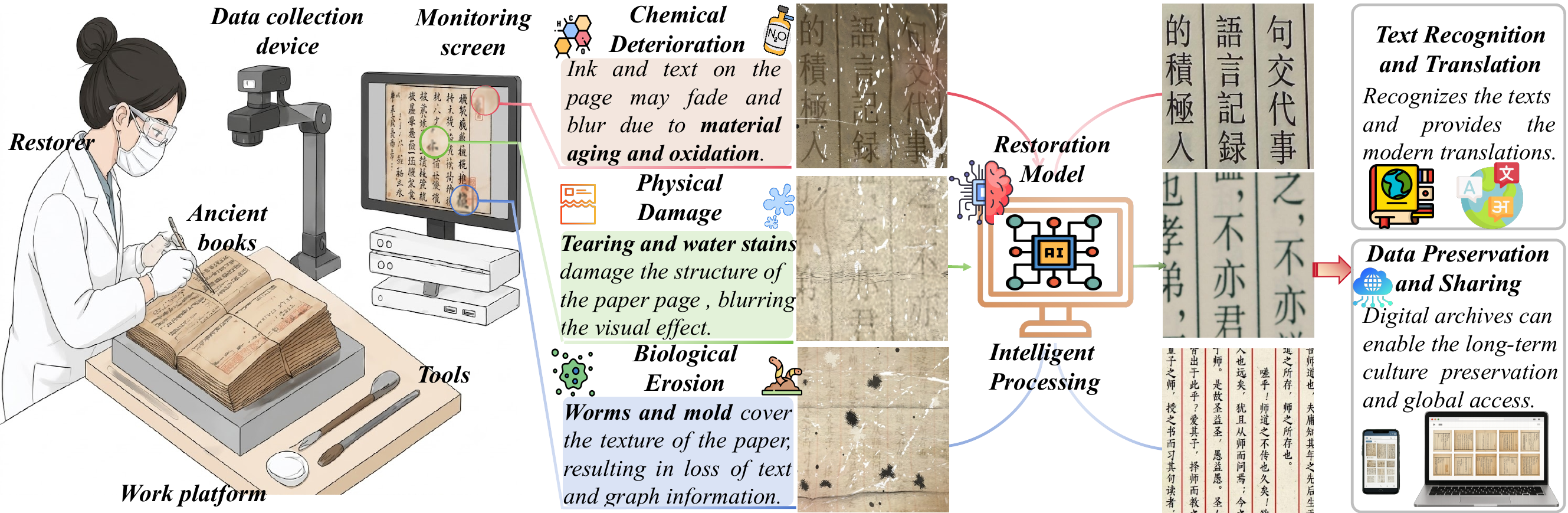}
	\vspace{-4mm}
	\caption{Overview of the ancient book image restoration process. Intelligent techniques are used to repair damage, followed by text recognition preservation and sharing.}
	\label{fig1}
	\vspace{-4mm}
\end{figure}

Ancient books preserve historical knowledge, cultural traditions, and material practices. During long-term preservation, paper and ink undergo chemical deterioration, physical damage, and biological erosion, resulting in discoloration, staining, deformation, mold, insect holes, and textual loss. Physical conservation remains indispensable for protecting original artifacts, but its cost and operational requirements limit its application to cultural heritage preservation. As illustrated in Figure~\ref{fig1}, digital image restoration provides a non-invasive complement by processing digitized pages without modifying the physical objects to improve the readability and accessibility of damaged digital surrogates. Unlike generic inscription restoration task, ancient-book image restoration recover the material appearance of the page while retaining text strokes, layout, and local structural details.

﻿

﻿

Learning-based image restoration methods have achieved strong results on generic image restoration using convolutional and Transformer backbones \cite{chen2022simple,zamir2020learning,wang2022uformer}, adaptive all-in-one models \cite{li2022all,potlapalli2023promptir,cui2025adair,tian2025degradation,yang2024language}, and diffusion-based methods \cite{xia2023diffir,zheng2024diffuir,luo2025visualinstructed}. However, their effectiveness depends on training data that reflect the degradation patterns of the target domain. For ancient books, collecting real clean--degraded pairs is rarely possible, making synthetic paired data essential for supervised restoration. Existing cultural-heritage benchmarks mainly target inscriptions, character-level reconstruction \cite{zhu2024reproducing,zhang2025reviving,yang2025predicting,zhu2026epiagent,li2024towards}. Their synthetic data are often organized around individual degradation types or generated by adding mask-based defects and local perturbations. Such constructions are insufficient for ancient-book image restoration, where paper ageing, staining, water-induced deformation, mold, insect damage, content loss, and blur frequently coexist and affect the same region. The appearance of a water-stained area, for example, depends on the aged paper and ink beneath it, while deformation can further alter nearby strokes and layout. Moreover, character- or region-centered data do not provide the page-level aligned supervision needed to restore damaged content while preserving the paper background, layout, and overall visual consistency of an ancient-book page.

To address the lack of paired supervision for composite ancient-book degradation, we introduce \textbf{AncBench}, a large-scale benchmark containing 100,379 aligned clean--degraded image pairs. Its core is a process-aware synthesis framework that models degradation as a state-dependent evolution of the page rather than an independent combination of predefined corruptions. Each degradation process operates on the page state produced by preceding processes, thereby preserving the dependencies among material ageing, geometric deformation, structural damage, and imaging quality. In particular, global material changes condition the appearance of subsequent local damage, while spatial transformations are consistently propagated across paper texture, text strokes, and damaged regions. By coupling degradation generation with transformation-aware alignment, AncBench produces composite observations while retaining reliable page-level supervision. This dependency-preserving formulation better approximates the formation of real ancient-book degradation and reduces the distribution gap between synthetic training data and practical restoration scenarios.
﻿

In addition to suitable data, ancient-book image restoration requires a model that can distinguish page-wide degradation from spatially concentrated damage. Existing methods \cite{zamir2022restormer,chen2022simple,xia2023diffir,li2022all,potlapalli2023promptir} are primarily designed for generic restoration tasks and typically model degradation globally or select from fixed degradation priors, which limits their ability to handle composite degradation. To address this limitation, we propose \textbf{AncRestore}, a degradation-aware visual-prior routing network. AncRestore begins with a Damage Representation Block (DRB), which encodes the degraded input at complementary levels. DRB produces a global degradation vector to describe the overall condition of the page, local degradation tokens to retain regional variation, a soft damage map to identify damage-sensitive areas, and a multi-scale damage-prior pyramid aligned with the restoration backbone. The Dynamic Expert Routing (DER) module then integrates multi-scale DINO visual priors \cite{simeoni2025dinov3} through expert-wise and region-wise routing. The former estimates the relevance of each expert to the current image, while the latter modulates its contribution across spatial locations. The routed priors condition the Half-quadratic splitting (HQS) restoration backbone \cite{he2013half}, achieving global recovery and reconstruction of damaged regions.

Overall, our contributions are summarized as follows:
\begin{itemize}
	\item We introduce AncBench, a large-scale paired benchmark with 100,379 pixel-aligned image pairs, alleviating the scarcity of high-quality paired data for ancient-book image restoration under composite degradation.
	﻿
	\item We propose AncRestore, a degradation-aware visual-prior routing network. Its DRB extracts global degradation and local damage representations, while DER uses expert-wise and region-wise routing to select and distribute multi-level visual priors for restoration. 
	﻿
	\item Extensive experiments show the effectiveness of AncBench and AncRestore in restoring degraded images while preserving text structures and paper appearance, facilitating data preservation and wider dissemination.
\end{itemize}

\section{Related Work}

\subsection{Datasets for Ancient Book Image Restoration}

Paired data are essential for supervised restoration, yet obtaining the original appearance of damaged ancient book is rarely feasible. Existing benchmarks therefore derive paired supervision from real collections and synthetic degradation. CIRI combines real Chinese inscription rubbings with synthesized intact--degraded pairs for inscription restoration \cite{zhu2024reproducing}. HDR28K formulates historical-document repair as original-appearance prediction and provides character-level annotations; its synthetic data comprise three separately constructed subsets for character missing, paper damage, and ink erosion \cite{yang2025predicting}. FPHDR extends the setting to complete historical-document pages, supplementing its images with character- and line-level locations, character annotations, and damage grades to support semantically guided restoration \cite{zhang2025reviving,pratikakis2019dibco}. Together, these datasets provide useful supervision for inscription and historical-document repair, but their scale and synthetic diversity remain limited for image restoration of ancient books. Their degradation models are typically factorized into a small number of isolated effects. For example, HDR28K relies primarily on randomly generated masks and region-specific corruption to construct its three degradation subsets, rather than modeling the co-occurrence and interaction of deterioration processes. Moreover, CIRI and HDR28K primarily target inscriptions or character-centered repair.
AncBench instead focuses on pixel-level reconstruction of full ancient-book pages under jointly applied chemical deterioration, physical damage and biological erosion.

\begin{figure*}[!t]
	\centering
	\begin{tabular}{cc}
		\hspace{-2mm}
		\includegraphics[width=0.41\linewidth]{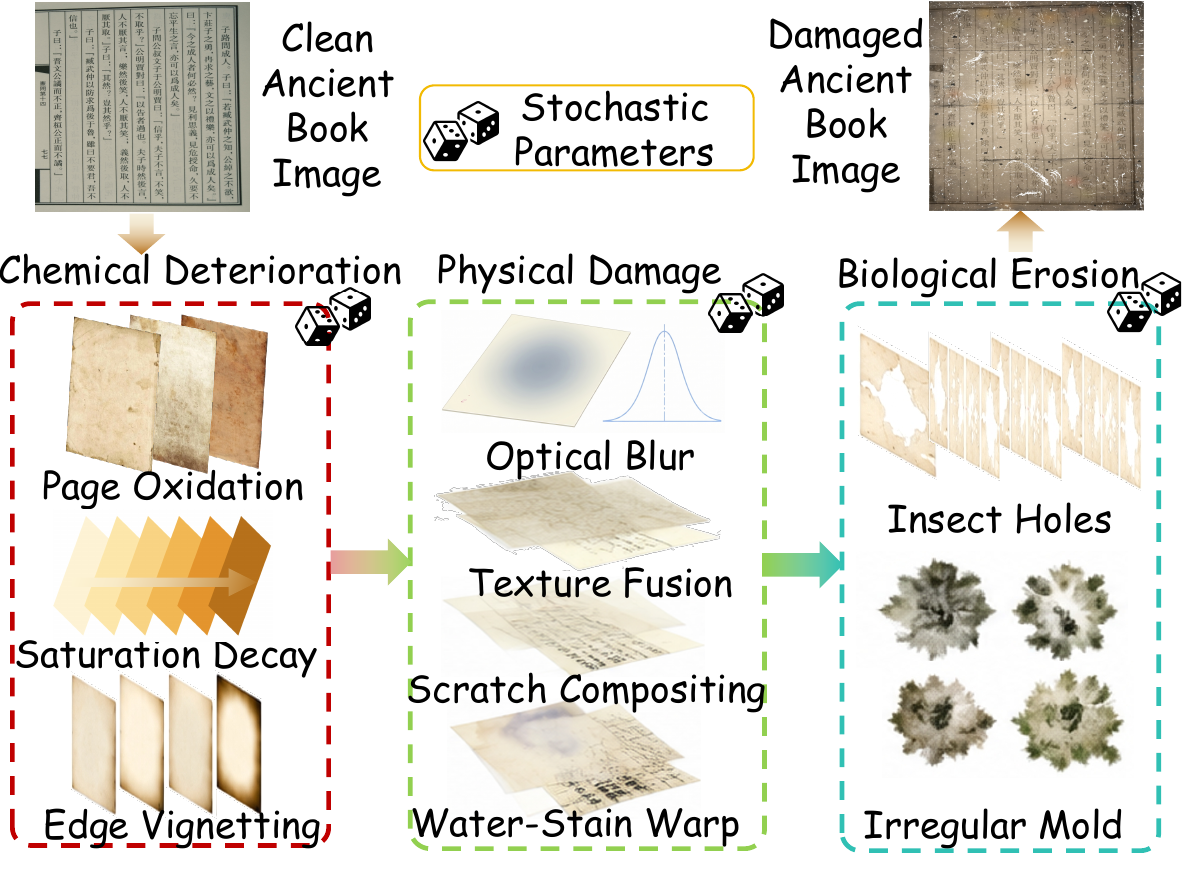} 
		\includegraphics[width=0.57\linewidth]{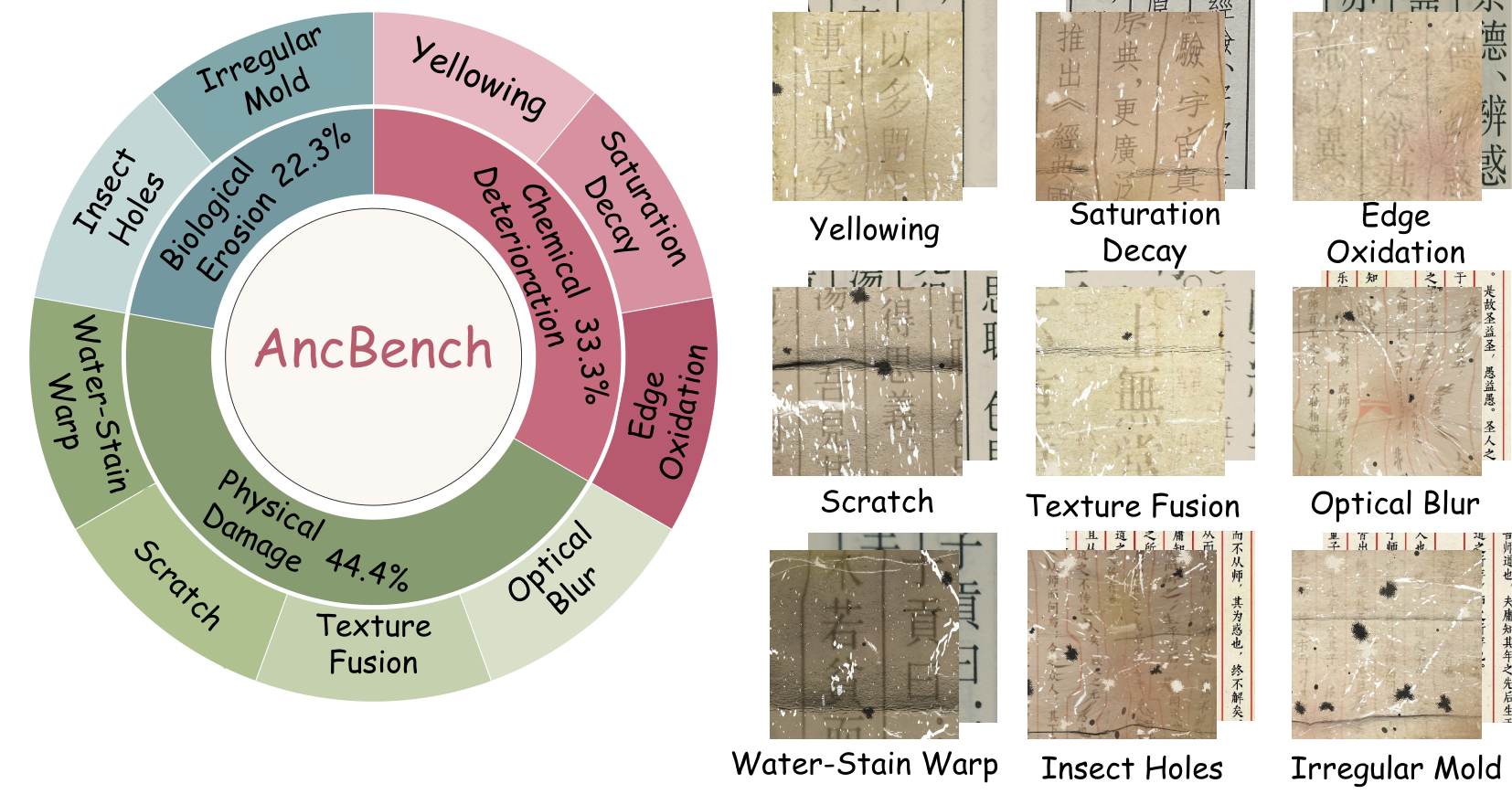} 
	\end{tabular}
	\caption{Construction and composition of the AncBench dataset. Clean ancient book images are synthetically degraded with chemical, physical, and biological damage using stochastic parameter; representative examples are shown on the right.} 
	\label{fig2} 
\end{figure*}

\subsection{Methods for Ancient Book Image Restoration}
﻿
General image-restoration methods have achieved strong performance on standard low-level tasks, including denoising, deblurring, and image enhancement, using Transformer-based architectures and diffusion models \cite{zamir2022restormer,chen2022simple,xia2023diffir}. All-in-one restoration methods further introduce degradation-conditioned processing to handle multiple corruption types within a single model \cite{li2022all,potlapalli2023promptir,tu2026dvanet,yang2024language}. These methods are primarily designed for generic degradation settings and do not explicitly model the joint variation of paper appearance, page geometry, text strokes, and imaging quality in ancient-book images.
Domain-specific methods incorporate document structure or textual priors. CIRI combines global and local diffusion restoration with inscription knowledge \cite{zhu2024reproducing}. DiffHDR uses semantic and spatial priors for damaged-document repair \cite{yang2025predicting}, while AutoHDR integrates OCR-based localization, contextual text prediction, and appearance reconstruction \cite{zhang2025reviving}. EpiAgent organizes multimodal analysis, historical knowledge, and specialized restoration tools within an agent-based framework for inscription restoration \cite{zhu2026epiagent}. Although effective in their intended settings, they are commonly trained and evaluated on a small set of degradation on inscription- and character-centered scenarios, leading to sub-optimal performance under the composite degradation patterns.

\section{Proposed Dataset}
\subsection{Data Collection}

We construct the clean-source collection from page images of representative Confucian and Daoist classics, primarily \textit{The Analects} and \textit{Su Shu}. We select high-resolution pages with complete layouts, legible character strokes, and no severe visible damage. The source pages are captured using a full-frame Canon EOS R6 Mark II camera under low-ISO settings. During acquisition, each page is placed on a planar surface, while controlled illumination, reflection suppression, and color calibration are applied to preserve fine stroke details and the original appearance of the paper. The captured images are stored in a lossless format and processed by geometric rectification, page-boundary cropping, resolution normalization, and reference-based color alignment.

\subsection{Coupled Degradation Synthesis}
\label{sec:degradation_synthesis}

AncBench represents degradation as a sequential composition of chemical deterioration, physical damage, and biological erosion. Given a clean page ($I_{\mathrm{GT}}$), the corresponding degraded image is generated as:
\begin{equation} I_{\mathrm{LQ}} = \mathcal{M} \circ \mathcal{W} \circ \mathcal{H} \circ \mathcal{S} \circ \mathcal{T} \circ \mathcal{V} \circ \mathcal{B} \circ \mathcal{C}(I_{\mathrm{GT}}), \label{eq:degradation_chain} \end{equation}
where ($\mathcal{C}$) models global chemical ageing, ($\mathcal{B}$) applies optical blur, and ($\mathcal{V}$) simulates edge oxidation. ($\mathcal{T}$) and ($\mathcal{S}$) denote paper-texture blending and scratch compositing, respectively, while ($\mathcal{H}$) synthesizes insect holes. ($\mathcal{W}$) introduces water stains together with local geometric deformation, and ($\mathcal{M}$) generates irregular mold patterns. Each operator acts on the output of the preceding stage. The operators are divided into chemical, physical, and biological damage families, and each synthesized LQ image includes degradation from all families.

﻿\noindent\textbf{Chemical Deterioration.}
Chemical ageing alters the overall color and luminance distribution of paper and ink. We simulate this process by introducing a warm color shift and reducing color saturation to reproduce paper yellowing and pigment fading. Ageing also causes non-uniform oxidation, with stronger discoloration and darkening frequently occurring near page boundaries. We model this spatial variation using an edge-oxidation operator, which can be defined as:
\begin{equation}
	\mathcal{V}(I)(p)
	=
	I(p)\cdot
	\max\left(
	v_{\min},
	1-\kappa\frac{d(p,c)}{d_{\max}}
	\right),
	\label{eq:edge_oxidation}
\end{equation}
where $c$ is the image center, $d(p,c)$ denotes the Euclidean distance between pixel $p$ and $c$, and $d_{\max}$ is the maximum distance from the center to the image boundary. The coefficient $\kappa$ controls the radial attenuation, while $v_{\min}$ specifies its lower bound. Pixels farther will produce darker boundaries.

\noindent\textbf{Physical Damage.}
Physical degradation includes optical blur, paper-texture variation, and water-induced deformation. Let $G_{\sigma}$ denote a Gaussian kernel, $T$ a sampled paper-texture image, and $(A_{\mathrm{s}},S)$ the alpha mask and color field of a scratch overlay. The first three effects are modeled as:
\begin{equation}
	\begin{aligned}
		\mathcal{B}(I) &= G_{\sigma} * I, \\
		\mathcal{T}(I)(p) &=
		(1-\alpha_{\mathrm{t}})I(p)
		+\alpha_{\mathrm{t}}T(p), \\
		\mathcal{S}(I)(p) &=
		\bigl(1-A_{\mathrm{s}}(p)\bigr)I(p)
		+A_{\mathrm{s}}(p)S(p),
	\end{aligned}
	\label{eq:physical_compositing}
\end{equation}
where $\alpha_{\mathrm{t}}$ controls the texture-blending strength. This formulation preserves the page content while introducing substrate variation and localized abrasion.
Water damage jointly changes local appearance and page geometry. For each sampled stain, we construct an opacity map $A_{\mathrm{w}}$, a brownish stain-color field $C_{\mathrm{w}}$, and a radial--tangential displacement field. The degraded pixel is given by:
\begin{equation}
	\begin{aligned}
		p' &= p+\Delta_{\mathrm{rad}}(p)+\Delta_{\mathrm{tan}}(p), \\
		\mathcal{W}(I)(p) &=
		\bigl(1-A_{\mathrm{w}}(p)\bigr)I(p')
		+A_{\mathrm{w}}(p)C_{\mathrm{w}}(p),
	\end{aligned}
	\label{eq:water_deformation}
\end{equation}
where $\Delta_{\mathrm{rad}}(p)$ and $\Delta_{\mathrm{tan}}(p)$ denote the radial and tangential displacement components around the stain center.

﻿

\begin{figure*}[!t]
	\centering
	% \vspace{-2mm}
	\begin{tabular}{cc}
		\hspace{-2mm}
		\includegraphics[width=0.98\linewidth]{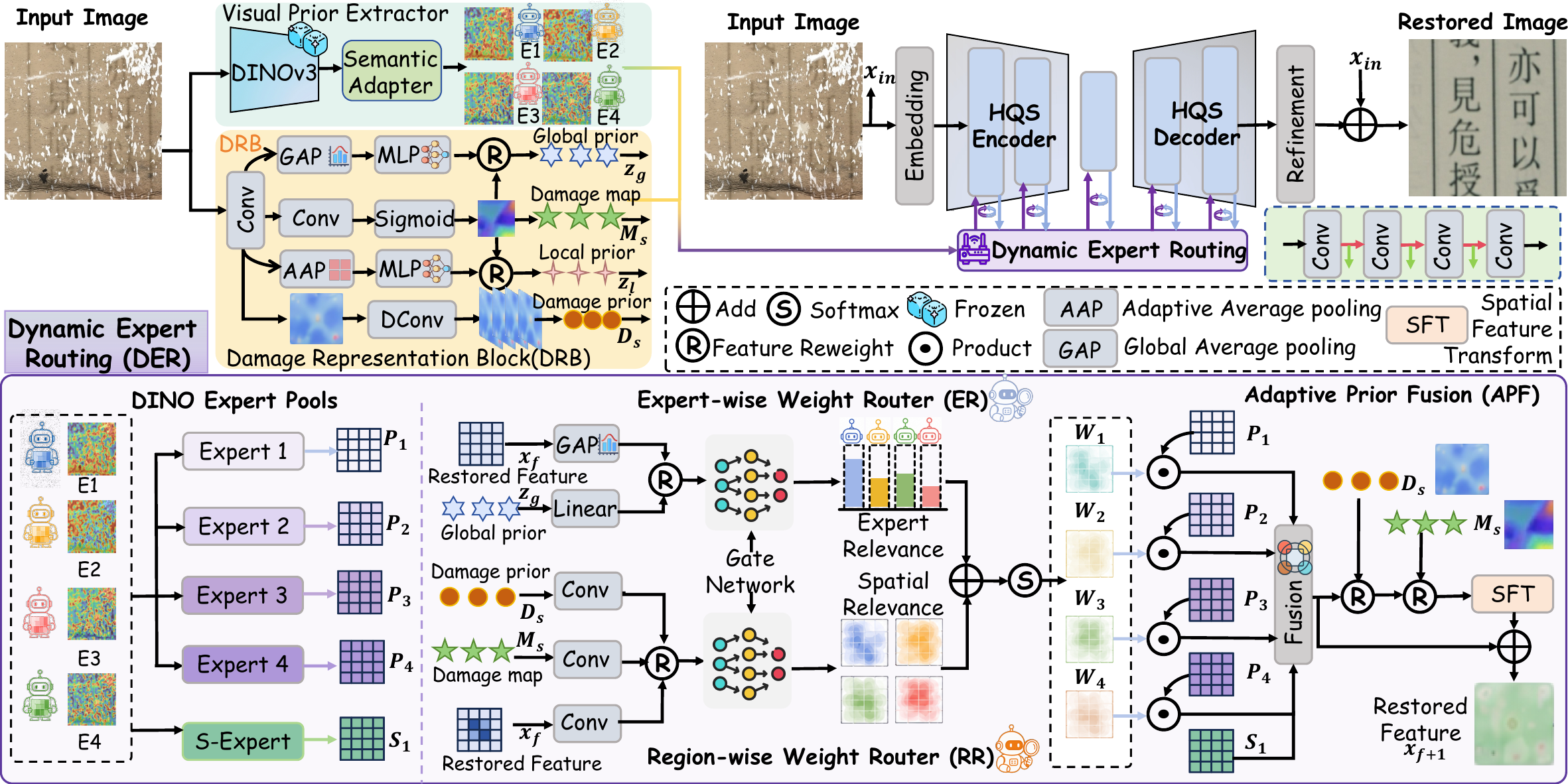} 
	\end{tabular}
	\caption{Overview of the proposed AncRestore. AncRestore combines degradation representation, HQS-based restoration backbone and dynamic expert routing to adaptively select and fuse multi-level visual priors for ancient-book image restoration. 
	} 
	\label{fig3} 
\end{figure*}

\noindent\textbf{Biological Erosion.}
The biological-damage family models insect-related loss and mold growth. We first sample $N_{\mathrm{h}}$ insect-hole regions $\{\Omega_j^{\mathrm{h}}\}_{j=1}^{N_{\mathrm{h}}}$ and construct their union mask:
\begin{equation}
	\begin{aligned}
		A_{\mathrm{h}}(p)
		&=
		\max_{1\leq j\leq N_{\mathrm{h}}}
		\mathbf{1}_{\Omega_j^{\mathrm{h}}}(p),
		\\
		\mathcal{H}(I)(p)
		&=
		\bigl(1-A_{\mathrm{h}}(p)\bigr)I(p)
		+A_{\mathrm{h}}(p)C_{\mathrm{h}},
		\label{eq:insect_holes}
	\end{aligned}
\end{equation}
where $\mathbf{1}{\Omega_j^{\mathrm{h}}}$ is the indicator function of the $j$-th hole region and $C{\mathrm{h}}$ denotes its dark interior color.

Mold clusters are generated with irregular boundaries instead of circular masks. For the $j$-th cluster centered at $c_j$, its boundary radius is defined in polar coordinates as:
\begin{equation}
	r_j(\theta)
	=
	r_{j,0}\bigl(1+\epsilon_j(\theta)\bigr),
	\quad
	\Omega_j^{\mathrm{m}}
	=
	\left\{
	p:\|p-c_j\|\leq r_j(\theta_p)
	\right\},
	\label{eq:mold_shape}
\end{equation}
where $r_{j,0}$ is the base radius and $\epsilon_j(\theta)$ combines perturbations over eight angular sectors with spatial noise. The resulting mold mask and compositing operation are:
\begin{equation}
	\begin{aligned}
		A_{\mathrm{m}}(p)
		&=
		1-\prod_{j=1}^{N_{\mathrm{m}}}
		\left(
		1-\alpha_j\mathbf{1}_{\Omega_j^{\mathrm{m}}}(p)
		\right), \\
		\mathcal{M}(I)(p)
		&=
		\bigl(1-A_{\mathrm{m}}(p)\bigr)I(p)
		+A_{\mathrm{m}}(p)C_{\mathrm{m}}(p),
	\end{aligned}
	\label{eq:mold_compositing}
\end{equation}
where $N_{\mathrm{m}}$ is the number of mold clusters, $\alpha_j$ is the opacity of the $j$-th cluster, and $C_{\mathrm{m}}(p)$ is its spatially varying color.

\subsection{Dataset Statistics}
\label{sec:dataset_statistics}
﻿
Figure~\ref{fig2} summarizes the composition and degradation coverage of AncBench. The released patch benchmark contains 100,379 aligned GT/LQ pairs at $512\times512$ resolution, including 99,879 training pairs and 500 test pairs. The center summarizes the nine degradation implemented in the synthesis pipeline. The corresponding proportions, $33.3\%$, $44.4\%$, and $22.3\%$, indicate the coverage of the implemented taxonomy. The right-hand examples isolate the visible effect of each operator on a clean page, including global material changes, local surface damage, water-induced distortion, and biological erosion. This construction produces paired data in which global paper appearance and localized structural damage co-occur within the same page.

\section{Method}
\label{sec:method}

\subsection{Overview}
\label{sec:method_overview}
Given a degraded ancient-book image $I_{\mathrm{in}}$, AncRestore reconstructs the restored image $I_{\mathrm{out}}$ by jointly modeling global material degradation and local text damage. As shown in Figure~\ref{fig3}, the framework comprises a frozen DINOv3-based visual-prior extractor, a Damage Representation Block (DRB), a Dynamic Expert Routing module (DER), and an HQS-based encoder--decoder backbone. Four intermediate DINOv3 features are projected by a semantic adapter to form a visual-expert pool. Meanwhile, DRB extracts global and local degradation representations, a spatial damage map, and multi-scale damage priors from the input image. DER then dynamically selects and spatially modulates the visual experts according to these degradation cues, and injects the fused priors into each restoration stage. Finally, the backbone predicts a residual image to obtain the restored output.
﻿

\subsection{Damage Representation Block}
\label{sec:drb}

Ancient-book degradations have both broad effects, such as material ageing and color change, and concentrated effects, such as stains, insect holes, or missing strokes. DRB is designed to describe these coupled effects without requiring damage-mask annotations. As shown in Figure~\ref{fig3}, a shallow convolution is applied to produce a shared feature map $F$ from $I_{\mathrm{in}}$. DRB then pools $F$ at global and local scales and maps the pooled features to the degradation representations:
\begin{equation}
	z_g^0=g_g\!\left(\operatorname{GAP}(F)\right),\qquad
	Z_l^0=g_l\!\left(\operatorname{AAP}_{2\times2}(F)\right),
	\label{eq:initial_degradation_representation}
\end{equation}
where $\operatorname{GAP}(\cdot)$ denotes global average pooling operation and $\operatorname{AAP}_{2\times2}(\cdot)$ denotes adaptive average pooling over a $2\times2$ grid. The MLPs $g_g(\cdot)$ and $g_l(\cdot)$ produce an image-level degradation vector $z_g^0$ and four local tokens $Z_l^0$, respectively. These two branches retain both the overall material condition and coarse differences across regions.

Global and regional pooling summarize degradation statistics but discard the precise spatial distribution of localized defects. To retain this information, DRB predicts a soft damage map from the shared feature:
\begin{equation}
	M_s(u)=\sigma\!\left(h_d(F)(u)\right)\in(0,1),
	\label{eq:soft_damage_map}
\end{equation}
where $h_d(\cdot)$ is a convolutional damage-prediction head and $u$ indexes a spatial location. The map $M_s$ enhances the pooled global and local features before they are added to the corresponding representations:
\begin{equation}
	\begin{aligned}
		z_g=z_g^0+\lambda_g\,g_g^d\!\left(\operatorname{GAP}(F\odot M_s)\right),\\
		Z_l=Z_l^0+\lambda_l\,g_l^d\!\left(\operatorname{AAP}_{2\times2}(F\odot M_s)\right),
	\end{aligned}
	\label{eq:damage_aware_representation}
\end{equation}
where $\odot$ denotes element-wise multiplication, $g_g^d(\cdot)$ and $g_l^d(\cdot)$ are MLPs, and $\lambda_g$ and $\lambda_l$ are learned gates. Thus, DRB emphasizes affected regions while retaining the original global and local representations.

The same soft map also initializes a multi-scale prior pyramid for DER. Specifically, DRB enhances the feature map with the estimated damage and progressively projects it to the four backbone resolutions:
\begin{equation}
	\begin{aligned}
		D_s^1=\pi_1\!\left(F\odot(1+M_s)\right),\\
		D_s^{r+1}=\pi_{r+1}\!\left(D_s^r\right),\quad r=1,2,3,
		\label{eq:damage_prior_pyramid}
	\end{aligned}
\end{equation}
where $\pi_r(\cdot)$ denotes the projection at level $r$. Thus, DRB provides a compact global description, regional damage cues, and scale-matched priors for subsequent restoration.

\subsection{Dynamic Expert Routing}
\label{sec:der}
The four adapted DINOv3 features serve as complementary visual experts, as different layers encode distinct structural and semantic cues. Let $E_i^t=\mathcal{R}_t(E_i)$ denote the $i$-th expert aligned to restoration stage $t$. DER constructs expert-specific priors and a shared prior as:
\begin{equation}
	P_i^t=p_i(E_i^t),\quad P_{\mathrm{sh}}^t=\frac{1}{N}\sum_{i=1}^{N}p_s(E_i^t),\quad N=4,
	\label{eq:expert_priors}
\end{equation}
where $p_i(\cdot)$ is expert-specific and $p_s(\cdot)$ is shared across experts. All experts remain available at each stage and are dynamically weighted by DER.

The scale-matched damage prior $D_s^t$ and damage map $M_s^t$ are projected into a joint routing context:
\begin{equation}
	Q_D^t=q_D(D_s^t),\quad Q_M^t=q_M(M_s^t),\quad C^t=q_C([Q_D^t,Q_M^t]),
	\label{eq:damage_routing_context}
\end{equation}
where $q_D(\cdot)$, $q_M(\cdot)$, and $q_C(\cdot)$ are convolution projections, and $[\cdot,\cdot]$ denotes channel-wise concatenation.

\paragraph{Expert-wise routing.}
ER assigns an image-level score to each expert from the restoration state and global degradation condition. The damage-enhanced feature is globally pooled and projected together with $z_g$:
\begin{equation}
	h_{\mathrm{E}}^t=\delta\!\left(W_x\operatorname{GAP}(x_f^t+\lambda_c C^t)+W_gz_g\right),\quad
	a^t=W_Eh_{\mathrm{E}}^t,
	\label{eq:expert_wise_routing}
\end{equation}
where $W_x$ and $W_g$ project the restoration and degradation representations, respectively, $W_E$ is the routing head, $\delta(\cdot)$ denotes the activation function, and $\lambda_c$ is a learnable gate. Each $a_i^t$ indicates the image-level compatibility of expert $i$ with the current degradation state.

\paragraph{Region-wise routing.}
Since image-level routing cannot resolve spatially varying damage, RR predicts an expert score from the restored feature, damage prior, and damage map:
\begin{equation}
	h_{\mathrm{R}}^t=\delta\!\left(W_{\mathrm{R},1}[x_f^t,Q_D^t,Q_M^t]\right),\qquad
	b^t=W_{\mathrm{R},2}h_{\mathrm{R}}^t,
	\label{eq:region_wise_routing}
\end{equation}
where $W_{\mathrm{R},1}$ and $W_{\mathrm{R},2}$ are convolutional layers. The value $b_i^t(u)$ denotes the local relevance of expert $i$ at location $u$.
The global and regional scores are normalized jointly:
\begin{equation}
	w_i^t(u)=
	\frac{\exp\!\left((a_i^t+b_i^t(u))/T\right)}
	{\sum_{j=1}^{N}\exp\!\left((a_j^t+b_j^t(u))/T\right)},
	\label{eq:routing_weights}
\end{equation}
where $T$ is the routing temperature. The global score $a_i^t$ is shared across spatial locations, while $b_i^t(u)$ provides a location-dependent correction.
The priors are aggregated as:
\begin{equation}
	P_{\mathrm{pr}}^t=\sum_{i=1}^{N}w_i^t\odot P_i^t.
	\label{eq:routed_private_prior}
\end{equation}
Adaptive Prior Fusion then combines the routed private prior, shared prior, and damage context:
\begin{equation}
	G^t=\sigma\!\left(g\!\left([x_f^t+\lambda_cC^t,P_{\mathrm{pr}}^t,P_{\mathrm{sh}}^t]\right)\right),
	\label{eq:private_shared_gate}
\end{equation}
\begin{equation}
	P_{\mathrm{fuse}}^t
	=
	G^t\odot P_{\mathrm{pr}}^t
	+
	(1-G^t)\odot P_{\mathrm{sh}}^t
	+
	\lambda_d C^t,
	\label{eq:prior_fusion}
\end{equation}
where $g(\cdot)$ is a convolutional fusion head, $G^t$ is a spatial gate, and $\lambda_d$ is learnable. A spatial feature transform maps $P_{\mathrm{fuse}}^t$ to scale and shift tensors $(\gamma^t,\beta^t)$, which modulate the restoration feature through a gated residual update:
% The fused prior is mapped to affine modulation parameters and injected into the feature:
\begin{equation}
	\widetilde{x}_f^t
	=
	x_f^t
	+
	\lambda_s
	\left(
	x_f^t\odot(1+\gamma^t)+\beta^t-x_f^t
	\right),
	\label{eq:sft_fusion}
\end{equation}
where $\psi(\cdot)$ is the SFT projection and $\lambda_s$ is a learnable gate.

\begin{table*}[t]\footnotesize
	\centering
	\setlength{\tabcolsep}{4.8pt}
	\renewcommand{\arraystretch}{1}
	\begin{tabular}{@{}llcccccccc@{}}
		\toprule
		\multirow{2}{*}{\textbf{Methods}} & \multirow{2}{*}{\textbf{Publication}} & \multicolumn{5}{c}{\textbf{Quality}} & \multicolumn{3}{c}{\textbf{Recognition}} \\
		\cmidrule(lr){3-7}\cmidrule(lr){8-10}
		&  & \textbf{PSNR$\uparrow$} & \textbf{SSIM$\uparrow$} & \textbf{LPIPS$\downarrow$} & \textbf{MUSIQ$\uparrow$} & \textbf{CLIP-IQA$\uparrow$} & \textbf{Rec-Acc.$\uparrow$} & \textbf{Precision $\uparrow$} & \textbf{F1 score $\uparrow$} \\
		\midrule
		MPRNet       & CVPR 2021    & 24.25 & 0.8979 & 0.1477 & 44.6083 & 0.3565 & 85.23\% & 74.60\% & 79.56\% \\
		Restormer    & CVPR 2022    & 23.53 & 0.8897 & 0.1825 & 40.7427 & 0.3304 & 76.73\% & 75.08\% & 75.90\% \\
		PromptIR     & NeurIPS 2023 & 23.63 & 0.8907 & 0.1764 & 41.0094 & 0.3370 & 76.82\% & 73.66\% & 75.21\% \\
		IR-SDE       & ICML 2023    & 17.57 & 0.8130 & 0.2338 & \textbf{51.3281} & 0.3516 & 67.66\% & 70.91\% & 69.25\% \\
		SFHformer    & ECCV 2024    & 24.26 & 0.8900 & 0.1765 & 41.9468 & 0.3651 & 81.59\% & 75.13\% & 78.23\% \\
		BioIR        & NeurIPS 2025 & 24.67 & 0.8967 & 0.1777 & 42.3489 & 0.3848 & 83.41\% & 75.57\% & 79.30\% \\
		AdaIR        & ICLR 2025    & 25.24 & \underline{0.9090} & \underline{0.1348} & 44.4679 & \underline{0.3899} & 86.54\% & \underline{76.59\%} & 81.26\% \\
		HOGformer    & AAAI 2026    & 24.63 & 0.8932 & 0.1617 & 41.9979 & 0.3015 & 82.06\% & 75.11\% & 78.43\% \\
		MIRAGE       & ICLR 2026    & 24.94 & 0.9010 & 0.1524 & 43.7932 & 0.3832 & \underline{88.50\%} & 76.28\% & \underline{81.94\%} \\
		LWTformer    & CVPR 2026    & \underline{25.37} & 0.9078 & 0.1477 & 44.6757 & 0.3836 & 86.92\% & 76.01\% & 81.10\% \\
		\midrule
		\textbf{AncRestore (Ours)} &-- & \textbf{26.70} & \textbf{0.9191} & \textbf{0.1346} & \underline{45.2042} & \textbf{0.3970} & \textbf{89.58\%} & \textbf{76.99\%} & \textbf{82.81\%} \\
		\bottomrule
	\end{tabular}%
	\vspace{-2mm}
	\caption{Quantitative comparison on AncBench. Best and second-best results are highlighted in \textbf{bold} and \underline{underlined}, respectively.}
	\label{tab1}
\end{table*}

\begin{figure*}[!t]
	\vspace{-2mm}
	\centering
	\begin{tabular}{cc}
		\hspace{-3mm}
		\includegraphics[width=0.995\linewidth]{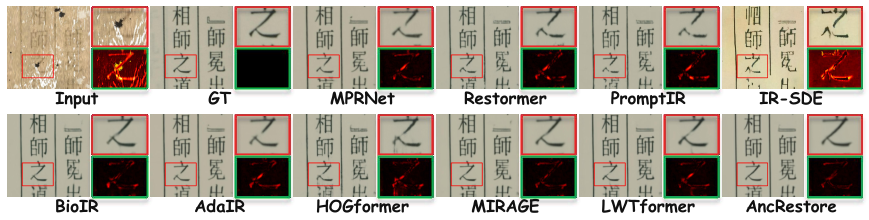}
		%		 \\
	\end{tabular}
	\vspace{-4mm}
	\caption{Qualitative comparison on AncBench. Red and green boxes show enlarged regions for detailed inspection. AncRestore more accurately reconstructs degraded character strokes while suppressing local artifacts and preserving the surrounding texture.}
	\label{fig5} 
	\vspace{-4mm}
\end{figure*}

\section{Experiments}
\label{sec:experiments}

\subsection{Experimental Setup}
\label{sec:experimental_setup}
\paragraph{Datasets and Metrics.}
We evaluate AncRestore on the proposed AncBench and two public benchmarks, HDR28K \cite{yang2025predicting} and HanInpaint. AncBench is used for training and evaluation under composite ancient-book degradation, while the two public datasets assess performance on historical-document repair and inpainting settings. Quantitative evaluation employs PSNR and SSIM for reconstruction fidelity and LPIPS for perceptual similarity.
To assess the preservation of textual content, we also report the accuracy of character recognizer on datasets with annotations.

\subsection{Comparison with Existing Methods}
Table~\ref{tab1} shows that AncRestore achieves the best overall performance on AncBench, reaching 26.70 dB PSNR, 0.9191 SSIM, and 0.1346 LPIPS. It also obtains the highest OCR accuracy and F1 score of 89.58\% and 82.81\%, respectively, confirming its effectiveness in both image restoration and text preservation. Although its MUSIQ score ranks second to IR-SDE, this reflects the different preferences of perceptual and reconstruction-oriented metrics.
%﻿
As shown in Figure~\ref{fig5}, AncRestore better preserves character boundaries and paper texture while reducing residual artifacts in damaged regions. These visual results are consistent with its quantitative gains.
﻿

\begin{figure*}[!t]
	\centering
	\begin{tabular}{c}
		\hspace{-3mm}\includegraphics[width=0.995\linewidth]{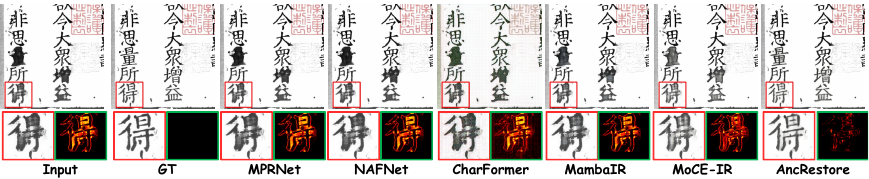} 
		%		\\
		%		\hspace{-3mm}\includegraphics[width=0.995\linewidth]{Figures/Fig6_4.pdf} 
	\end{tabular}
	\vspace{-3mm}
	\caption{Qualitative comparison on HDR28K. Red and green boxes show enlarged regions for detailed inspection. AncRestore more accurately reconstructs degraded character strokes while suppressing local artifacts and preserving the surrounding texture.
	} 
	\label{fig6_3} 
	\vspace{-4mm}
\end{figure*}

We further compare AncRestore with the representative restoration methods on the public HDR28K and HanInpaint datasets, including MPRNet, LWTformer, NAFNet~\cite{chen2022simple}, CharFormer~\cite{tay2022charformer}, MambaIR~\cite{guo2024mambair}, and MoCE-IR~\cite{zamfir2025moceir}.
﻿
As reported in Table~\ref{tab2}, AncRestore achieves strong generalization across different degradation types. It obtains the best PSNR on all three HDR28K subsets, including 21.60 dB for character missing, 23.86 dB for paper damage, and 28.89 dB for ink erosion. It also achieves the best SSIM and LPIPS on paper damage. On HanInpaint, AncRestore ranks first in SSIM and LPIPS and second in PSNR by only 0.06 dB. Although several perceptual metrics favor other methods under character missing and ink erosion, the overall results demonstrate robust reconstruction performance across external benchmarks. Figure~\ref{fig6_3} further shows that AncRestore better preserves character contours and page structures.

﻿\begin{table*}[t]
	\centering
	
	\setlength{\tabcolsep}{2.1pt}
	\renewcommand{\arraystretch}{1.2}
	\resizebox{\textwidth}{!}{%
		\begin{tabular}{llcccccccccccccccc}
			\toprule
			\multirow{3}{*}{\textbf{Methods}} & \multirow{3}{*}{\textbf{Publication}}
			& \multicolumn{12}{c}{\textbf{HDR28K}} & \multicolumn{4}{c}{\multirow{2}{*}{\textbf{HanInpaint}}} \\
			\cmidrule(lr){3-14}
			& & \multicolumn{4}{c}{\textbf{Character-missing}}
			& \multicolumn{4}{c}{\textbf{Paper-damage}}
			& \multicolumn{4}{c}{\textbf{Ink-erosion}}
			& & & \\
			\cmidrule(lr){3-6}\cmidrule(lr){7-10}\cmidrule(lr){11-14}\cmidrule(l){15-18}
			& &\textbf{ PSNR}$\uparrow$ &\textbf{ SSIM}$\uparrow$ & \textbf{LPIPS}$\downarrow$ & \textbf{MUSIQ}$\uparrow$
			&\textbf{ PSNR}$\uparrow$ &\textbf{ SSIM}$\uparrow$ & \textbf{LPIPS}$\downarrow$ & \textbf{MUSIQ}$\uparrow$
			&\textbf{ PSNR}$\uparrow$ &\textbf{ SSIM}$\uparrow$ & \textbf{LPIPS}$\downarrow$ & \textbf{MUSIQ}$\uparrow$
			&\textbf{ PSNR}$\uparrow$ &\textbf{ SSIM}$\uparrow$ & \textbf{LPIPS}$\downarrow$ & \textbf{MUSIQ}$\uparrow$ \\
			\midrule
			MPRNet     & CVPR 2021 & 20.51 & 0.9012 & 0.1350 & 56.5054 & 21.85 & \underline{0.9266} & \underline{0.1093} & 55.9959 & 27.16 & 0.9574 & \textbf{0.0376} & \textbf{55.1148} & 28.94 & 0.9675 & 0.0480 & 59.5651 \\
			NAFNet     & ECCV 2022 & 20.01 & 0.8976 & \underline{0.0912} & 53.7241 & 15.43 & 0.8869 & 0.1514 & 54.5393 & 23.19 & 0.9358 & 0.0600 & 54.9008 & \textbf{31.72} & \underline{0.9737} & \underline{0.0363} & \textbf{60.4726} \\
			CharFormer & ICLR 2022 & 15.03 & 0.4247 & 0.5918 & 41.9007 & 16.90 & 0.8224 & 0.1764 & \textbf{59.0359} & 18.41 & 0.6253 & 0.3021 & 50.8800 & 24.66 & 0.9006 & 0.1090 & 58.5199 \\
			MambaIR       & ECCV 2024 & 20.60 & \underline{0.9016} & 0.1321 & 55.1465 & 21.24 & 0.9193 & 0.1154 & 55.2657 & 23.29 & 0.9354 & 0.0594 & \underline{55.0047} & 28.71 & 0.9666 & 0.0447 & 59.1885 \\
			MoCE-IR    & CVPR 2025 & 20.01 & 0.8976 & \textbf{0.0911} & 53.7241 & 21.65 & 0.9251 & 0.1109 & 55.2098 & 23.45 & 0.9357 & 0.0586 & 54.9315 & 30.75 & 0.9704 & 0.0417 & \underline{60.3114} \\
			LWTformer  & CVPR 2026 & \underline{21.29} & \textbf{0.9027} & 0.1502 & \underline{57.8135} & \underline{22.23} & 0.9254 & 0.1119 & 56.6874 & \underline{28.43} & \underline{0.9622} & 0.0428 & 54.8465 & 30.19 & 0.9687 & 0.0426 & 60.1679 \\
			\midrule
			\textbf{AncRestore (Ours)} & --        & \textbf{21.60} & 0.8986 & 0.1544 & \textbf{58.7872} & \textbf{23.86} & \textbf{0.9292} & \textbf{0.1072} & \underline{58.9434} & \textbf{28.89} & \textbf{0.9628} & \underline{0.0423} & 54.9178 & \underline{31.66} & \textbf{0.9751} & \textbf{0.0349} & 60.2499 \\
			\bottomrule
	\end{tabular}}
	\vspace{-2mm}
	\caption{Quantitative comparison on HDR28K and HanInpaint. Higher PSNR, SSIM, and MUSIQ values are better; lower LPIPS values are better. The best and second-best results are highlighted in \textbf{bold} and \underline{underline}, respectively.}
	\label{tab2}
	\vspace{-2mm}
\end{table*}

\begin{figure}[!t]
	\centering
	\begin{tabular}{cc}
		\hspace{-3mm}
		\includegraphics[width=0.98\linewidth]{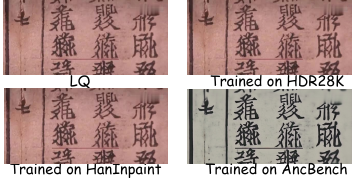} 
	\end{tabular}
	\vspace{-4mm}
	\caption{Real-world restoration results, which trained on different synthetic ancient book  image restoration datasets.
		\vspace{-4mm}
	} 
	\label{fig7} 
\end{figure}

\begin{table}[t]\footnotesize
	\centering
	\setlength{\tabcolsep}{4pt}
	\renewcommand{\arraystretch}{1.12}
	\begin{tabular}{@{}lccc ccc@{}}
		\toprule
		\multirow{2}{*}{\textbf{Variants}} & \multicolumn{2}{c}{\textbf{Components}} & \multicolumn{3}{c}{\textbf{Quality metrics}} \\
		\cmidrule(lr){2-3}\cmidrule(lr){4-6}
		& \textbf{DER} & \textbf{DRB} & \textbf{PSNR}$\uparrow$ & \textbf{LPIPS}$\downarrow$ & \textbf{MUSIQ}$\uparrow$ \\
		\midrule
		Baseline                & \ding{55}  &\ding{55}     & 25.56 & 0.1547 & 44.9086 \\
		Baseline+DRB          & \ding{55}   & \ding{51}   & 25.61 & 0.1553 & 43.9131 \\
		Baseline+DER                &\ding{51} & \ding{55}   & 25.80 & 0.1558 & 44.1917 \\
		\textbf{AncRestore}     & \ding{51} & \ding{51}  & \textbf{26.70} & \textbf{0.1346} & \textbf{45.2042} \\
		\bottomrule
	\end{tabular}%
	\vspace{-3mm}
	\caption{ Ablation studies on the effectiveness of the DER and DRB components in AncRestore.}
	\label{tab3}
	\vspace{-4mm}
\end{table}

\begin{table}[t]\footnotesize
	\centering
	\setlength{\tabcolsep}{4.2pt}
	\renewcommand{\arraystretch}{1}
	\begin{tabular}{@{}llccc@{}}
		\toprule
		\multirow{2}{*}{ \textbf{Components}} &\multirow{2}{*}{\textbf{Setting}}  & \multicolumn{3}{c}{\textbf{Quality metrics}} \\
		\cmidrule(lr){3-5}
		& & \textbf{PSNR}$\uparrow$ & \textbf{LPIPS}$\downarrow$ & \textbf{CLIP-IQA}$\uparrow$ \\
		\midrule
		\multirow{3}{*}{Routing} & RR only & 26.59 & 0.1485& 0.3825 \\
		& ER only     & 25.59 & 0.1568 & 0.3911 \\
		\midrule
		\multirow{3}{*}{No. of P-Experts} & 1 & 26.01 & 0.1540 & 0.3874 \\
		& 2 & 25.92 & 0.1547 & 0.3915 \\
		& 3 & 26.55 & 0.1554 & 0.3922 \\
		\midrule
		\textbf{AncRestore}  &\textbf{Full} & \textbf{26.70} & \textbf{0.1346} & \textbf{0.3970} \\
		\bottomrule
	\end{tabular}%
	\vspace{-3mm}
	\caption{Ablations on the prior routing strategies and the numbers of private experts in DER.}
	\label{tab4}
	\vspace{-4mm}
\end{table}

\subsection{ Analysis and Discussion}
\paragraph{Generalization ability of the AncBench.}
To evaluate the effectiveness of AncBench, we train the same restoration model separately on HDR28K, HanInpaint, and AncBench, and test them on an identical degraded input. As shown in the Figure~\ref{fig7}, the models trained on HDR28K and HanInpaint leave noticeable color distortion and residual degradation, whereas the model trained on AncBench recovers a cleaner paper appearance and clearer character strokes with fewer artifacts. This comparison indicates that the composite degradation patterns in AncBench better reflect the restoration requirements of ancient-book images, enabling the trained model to learn more transferable degradation representations and achieve stronger cross-domain restoration performance.
﻿
\paragraph{Effectiveness of DRB and DER.}
As shown in Table~\ref{tab3}, using DER or DRB alone provides limited and inconsistent improvements over the baseline. DER increases PSNR from 25.56 to 25.80 dB, whereas DRB yields a marginal gain to 25.61 dB. When both components are jointly employed, AncRestore achieves 26.70 dB PSNR, 0.1346 LPIPS, and 45.2042 MUSIQ, outperforming the baseline by 1.14 dB, 0.0201, and 0.2956, respectively. These results indicate that degradation representation and dynamic expert routing are complementary: DRB supplies degradation-aware guidance, while DER uses this information to select suitable priors.
﻿
\paragraph{Ablations on the various settings of DER.}
Table~\ref{tab4} further evaluates the routing design and the number of private experts. RR alone provides stronger reconstruction performance than ER alone, whereas ER achieves a higher CLIP-IQA score, suggesting that regional and global routing capture complementary spatial and semantic information. The complete routing strategy obtains the best results across all three metrics. Similarly, reducing the number of private experts leads to inferior performance, while the full configuration reaches 26.70 dB PSNR, 0.1346 LPIPS, and 0.3970 CLIP-IQA, confirming the benefit of combining global–regional routing with diverse expert priors.
﻿
﻿
\section{Conclusion}
This work establishes a robust benchmark and restoration framework for ancient-book images affected by complex and spatially varying degradations. AncBench contains 100K aligned clean--degraded image pairs and simulates diverse degradation factors, including material ageing, paper texture variation, blur, staining, geometric distortion, and local damage. Built upon this benchmark, AncRestore extracts global and local degradation representations and dynamically selects suitable visual priors through expert-wise and region-wise routing. This design enables the model to restore the overall page appearance while preserving character structures and local text details. Experiments on AncBench and public datasets demonstrate that AncRestore consistently improves restoration quality and text readability under composite degradations. These results confirm the effectiveness of AncBench and show the potential of AncRestore for ancient book digitization and cultural heritage preservation.

\bibliography{aaai2027}

@inbook{chen2022simple,
	author    = {Chen, Liangyu and Chu, Xiaojie and Zhang, Xiangyu and Sun, Jian},
	title     = {Simple Baselines for Image Restoration},
	booktitle = {ECCV},
	pages     = {17--33},
	year      = {2022}
}

@inproceedings{cui2025adair,
	author    = {Cui, Yuning and Zamir, Syed Waqas and Khan, Salman and Knoll, Alois and Shah, Mubarak and Khan, Fahad Shahbaz},
	title     = {AdaIR: Adaptive All-in-One Image Restoration via Frequency Mining and Modulation},
	booktitle = {ICLR},
	year      = {2025}
}

@inproceedings{guo2024mambair,
	author    = {Guo, Hang and Li, Jinmin and Dai, Tao and Ouyang, Zhihao and Ren, Xudong and Xia, Shu-Tao},
	title     = {MambaIR: A Simple Baseline for Image Restoration with State-Space Model},
	booktitle = {ECCV},
	year      = {2024},
	pages     = {222--241}
}

@inproceedings{li2022all,
	author    = {Li, Boyun and Liu, Xiao and Hu, Peng and Wu, Zhongqin and Lv, Jiancheng and Peng, Xi},
	title     = {All-In-One Image Restoration for Unknown Corruption},
	booktitle = {CVPR},
	pages     = {17431--17441},
	year      = {2022}
}

@inproceedings{li2024towards,
	title     = {Towards Automated Chinese Ancient Character Restoration: A Diffusion-Based Method with a New Dataset},
	author    = {Li, Haolong and Du, Chenghao and Jiang, Ziheng and Zhang, Yifan and Ma, Jiawei and Ye, Chen},
	booktitle = {AAAI},
	volume    = {38},
	number    = {4},
	pages     = {3073--3081},
	year      = {2024}
}

@inproceedings{luo2025visualinstructed,
	title     = {Visual-Instructed Degradation Diffusion for All-in-One Image Restoration},
	author    = {Luo, Wenyang and Qin, Haina and Chen, Zewen and Wang, Libin and Zheng, Dandan and Li, Yuming and Liu, Yufan and Li, Bing and Hu, Weiming},
	booktitle = {CVPR},
	year      = {2025},
	pages     = {12764--12777}
}

@inproceedings{potlapalli2023promptir,
	author    = {Potlapalli, Vaishnav and Zamir, Syed Waqas and Khan, Salman and Khan, Fahad Shahbaz},
	title     = {PromptIR: Prompting for All-In-One Image Restoration},
	booktitle = {NeurIPS},
	volume    = {36},
	pages     = {71275--71293},
	year      = {2023}
}

@inproceedings{pratikakis2019dibco,
	title     = {ICDAR 2019 Competition on Document Image Binarization (DIBCO 2019)},
	author    = {Pratikakis, Ioannis and Zagoris, Konstantinos and Karagiannis, Xenofon and Tsochatzidis, Lazaros and Mondal, Tanmoy and Marthot-Santaniello, Isabelle},
	booktitle = {ICDAR},
	year      = {2019},
	pages     = {1547--1556}
}

@article{he2013half,
	title={Half-quadratic-based iterative minimization for robust sparse representation},
	author={He, Ran and Zheng, Wei-Shi and Tan, Tieniu and Sun, Zhenan},
	journal={IEEE TPAMI},
	volume={36},
	number={2},
	pages={261--275},
	year={2013}
}

@misc{simeoni2025dinov3,
	title={{DINOv3}},
	author={Sim{\'e}oni, Oriane and Vo, Huy V. and Seitzer, Maximilian and Baldassarre, Federico and Oquab, Maxime and Jose, Cijo and Khalidov, Vasil and Szafraniec, Marc and Yi, Seungeun and Ramamonjisoa, Micha{\"e}l and Massa, Francisco and Haziza, Daniel and Wehrstedt, Luca and Wang, Jianyuan and Darcet, Timoth{\'e}e and Moutakanni, Th{\'e}o and Sentana, Leonel and Roberts, Claire and Vedaldi, Andrea and Tolan, Jamie and Brandt, John and Couprie, Camille and Mairal, Julien and J{\'e}gou, Herv{\'e} and Labatut, Patrick and Bojanowski, Piotr},
	year={2025},
	eprint={2508.10104},
	archivePrefix={arXiv},
}

@inproceedings{tay2022charformer,
	author    = {Tay, Yi and Tran, Vinh Q. and Ruder, Sebastian and Gupta, Jai and Chung, Hyung Won and Bahri, Dara and Qin, Zhen and Baumgartner, Simon and Yu, Cong and Metzler, Donald},
	title     = {Charformer: Fast Character Transformers via Gradient-based Subword Tokenization},
	booktitle = {ICLR},
	year      = {2022}
}

@inproceedings{tian2025degradation,
	title     = {Degradation-Aware Feature Perturbation for All-in-One Image Restoration},
	author    = {Tian, Xiangpeng and Liao, Xiangyu and Liu, Xiao and Li, Meng and Ren, Chao},
	booktitle = {CVPR},
	year      = {2025},
	pages     = {28165--28175}
}

@article{tu2026dvanet,
	title   = {DVANet: Degradation-Aware Visual-Prior Alignment Network for Image Restoration},
	author  = {Tu, Yanjie and Yan, Qingsen and Niu, Axi and Hu, Tao and Zhang, Haokui and Zhou, Jiantao},
	journal = {arXiv preprint arXiv:2606.19097},
	year    = {2026}
}

@inproceedings{wang2022uformer,
	title     = {Uformer: A General U-Shaped Transformer for Image Restoration},
	author    = {Wang, Zhendong and Cun, Xiaodong and Bao, Jianmin and Zhou, Wengang and Liu, Jianzhuang and Li, Houqiang},
	booktitle = {CVPR},
	year      = {2022},
	pages     = {17683--17693}
}

@inproceedings{xia2023diffir,
	author    = {Xia, Bin and Zhang, Yulun and Wang, Shiyin and Wang, Yitong and Wu, Xinglong and Tian, Yapeng and Yang, Wenming and Van Gool, Luc},
	title     = {DiffIR: Efficient Diffusion Model for Image Restoration},
	booktitle = {ICCV},
	pages     = {13049--13059},
	year      = {2023}
}

@inproceedings{yang2024language,
	title     = {Language-Driven All-in-One Adverse Weather Removal},
	author    = {Yang, Hao and Pan, Liyuan and Yang, Yan and Liang, Wei},
	booktitle = {CVPR},
	year      = {2024},
	pages     = {24902--24912}
}

@inproceedings{yang2025predicting,
	author    = {Yang, Zhenhua and Peng, Dezhi and Shi, Yongxin and Zhang, Yuyi and Liu, Chongyu and Jin, Lianwen},
	title     = {Predicting the Original Appearance of Damaged Historical Documents},
	booktitle = {AAAI},
	volume    = {39},
	number    = {9},
	pages     = {9382--9390},
	year      = {2025}
}

@inproceedings{zamfir2025moceir,
	author    = {Zamfir, Eduard and Wu, Zongwei and Mehta, Nancy and Tan, Yuedong and Paudel, Danda Pani and Zhang, Yulun and Timofte, Radu},
	title     = {Complexity Experts are Task-Discriminative Learners for Any Image Restoration},
	booktitle = {CVPR},
	year      = {2025}
}

@inproceedings{zamir2020learning,
	title     = {Learning Enriched Features for Real Image Restoration and Enhancement},
	author    = {Zamir, Syed Waqas and Arora, Aditya and Khan, Salman and Hayat, Munawar and Khan, Fahad Shahbaz and Yang, Ming-Hsuan and Shao, Ling},
	booktitle = {ECCV},
	year      = {2020},
	pages     = {492--511}
}

@inproceedings{zamir2022restormer,
	title     = {Restormer: Efficient Transformer for High-Resolution Image Restoration},
	author    = {Zamir, Syed Waqas and Arora, Aditya and Khan, Salman and Hayat, Munawar and Khan, Fahad Shahbaz and Yang, Ming-Hsuan},
	booktitle = {CVPR},
	year      = {2022},
	pages     = {5728--5739}
}

@inproceedings{zhang2025reviving,
	author    = {Zhang, Yuyi and Zhang, Peirong and Yang, Zhenhua and Yan, Pengyu and Shi, Yongxin and Liu, Pengwei and Guo, Fengjun and Jin, Lianwen},
	title     = {Reviving Cultural Heritage: A Novel Approach for Comprehensive Historical Document Restoration},
	booktitle = {ACL},
	pages     = {28876--28892},
	year      = {2025}
}

@inproceedings{zheng2024diffuir,
	author    = {Zheng, Dian and Wu, Xiao-Ming and Yang, Shuzhou and Zhang, Jian and Hu, Jian-Fang and Zheng, Wei-Shi},
	title     = {Selective Hourglass Mapping for Universal Image Restoration Based on Diffusion Model},
	booktitle = {CVPR},
	pages     = {25445--25455},
	year      = {2024}
}

@inproceedings{zhu2024reproducing,
	author    = {Zhu, Shipeng and Xue, Hui and Nie, Na and Zhu, Chenjie and Liu, Haiyue and Fang, Pengfei},
	title     = {Reproducing the Past: A Dataset for Benchmarking Inscription Restoration},
	booktitle = {ACM MM},
	pages     = {7714--7723},
	year      = {2024}
}

@inproceedings{zhu2026epiagent,
	title     = {EpiAgent: An Agent-Centric System for Ancient Inscription Restoration},
	author    = {Zhu, Shipeng and Chen, Ang and Nie, Na and Fang, Pengfei and Zhang, Min-Ling and Xue, Hui},
	booktitle = {CVPR},
	year      = {2026},
	pages     = {39304--39313}
}


\end{document}